# Implementation of a Hierarchical fuzzy controller for a biped robot.

**Abdallah Zaidi[1], Nizar Rokbani[1,2] and Adel.M. Alimi[1]**

[1] *REGIM Lab: Research Groups on Intelligent Machines Lab*
*National Engineering School of Sfax, University of Sfax,*
*BP w, Route de Soukra, Sfax, TUNISIA.*

[2] *Institute of Applied Sciences and technology of Sousse,*
*University of Sousse, TUNISIA.*

{abdallah.zaidi, nizar.rokbani, Adel.m.Alimi}@ieee.org

**Abstract.** In this paper the design of a control system for a biped robot is described. Control is specified for a walk cycle of the robot. The implementation of the control system was done on Matlab Simulink. In this paper a hierarchical fuzzy logic controller (HFLC) is proposed to control a planar biped walk. The HFLC design is bio-inspired from human locomotion system. The proposed method is applied to control five links planar biped into free area and without obstacles.

## 1 Introduction

The area of robotics has become more and more interesting in order to develop humane like machines that are able to perform tasks that previously have been possible only for humans to do[1, 2, 4]. In biped robotics intelligent techniques are investigated for walking gait generation [3, 5, 6, 8] and control. Gaits generators could be based on Inverse kinematics [12, 14, 20] or on solving the dynamics equation of motion. The complexity of the structure of the biped model requires more refined and delicate to obtain. It seems advantageous to use the techniques in intelligent control strategy for designing efficient and robust controllers [3, 8].

In this work we have reviewed the motivation, current state and future challenges of hierarchical fuzzy systems [19]. In [19], authors proposed an architecture of a hierarchical fuzzy controller that is not full. In conventional controllers based on fuzzy logic, the calculation complexity increases with the dimensions of the variables system. The proposed hierarchical fuzzy controller (HFLC) was implemented to reduce the number of rules to a linear function of variables system. However, the use of hierarchical fuzzy logic controllers raises new issues in the automatic design of controllers, namely the coordination of outputs of sub-controllers at lower levels of the hierarchy.

In this paper, the choice of a hierarchical fuzzy controller type as well as the demonstration of the hierarchical aspect of the biped robot itself are detailed. The reminder of the paper is organized as follows. In Section 2, the hierarchical fuzzy logic systems are introduced. Then a description of the biped robot used in our work is given in Section 3. The structure of the proposed hierarchical fuzzy controller is highlighted and discussed in Sections 4. Discussion and conclusion are explained in Section 5.

.

## 2 Hierarchical fuzzy systems

The hierarchical fuzzy systems have created summary to address a major problem suffered by standard fuzzy systems multi variables is actually the exponential progress in the number of rules based on the number of variables entered in the system. This problem is impractical for the purposes of its implementation for the size of the basic rules [15]. Several methods have been proposed and are based on the construction of multi variables functions from single variable function, so that all these methods have built their foundation on the idea of transforming the problem that has high dimension to the lower dimensions such as the basis of hierarchical fuzzy systems.

### *2.1 Hierarchical fuzzy systems of Raju*

The formulation of Raju was the basis for the design of the hierarchy Wang. Indeed, Zeng Ke gives the formulation of the basic rules of a hierarchical fuzzy system designed by Raju, this formulation is as follows: [18]

For Level 1, the base is described as follows:

**IF** $x_1$ **is** $A_{j_{1,1}}$ and $x_2$ **is** $A_{j_1,2}$ *and ... and* $x_{n1}$ **is** $A_{j_1,n_1}$ *then* $y_1$ *is* $b_1$

For level where i> 1 the rules are as follows:

**IF** $x_{N1+1}$ **is** $A_{j_{1,1}}$ *and* $x_2$ **is** $A_{j_1,2}$ *and ... and* $x_{N1+nj}$ **is** $A_{j_1,n_1}$ *and* $y_{i-1}$ *is* $b_{i-1}$ *then* $y_i$ *is* $b_i$

$\mathbf{N_i} = \sum_{k=1}^{i-1} n_k \leq \mathbf{n}$, $\mathbf{n_k}$ is the number of variables to enter the $k^{th}$ level, implied that $y_{i-1}$ are taken as variables and intermediate variables not enter the $i^{th}$ level.

The sufficient condition for a universal approximator is hierarchical fuzzy system. This system contained, in each level in the global system, universal approximators [18].

### *2.2 Joo hierarchical fuzzy systems*

The proposed hierarchy has the following structure (fig. 1):

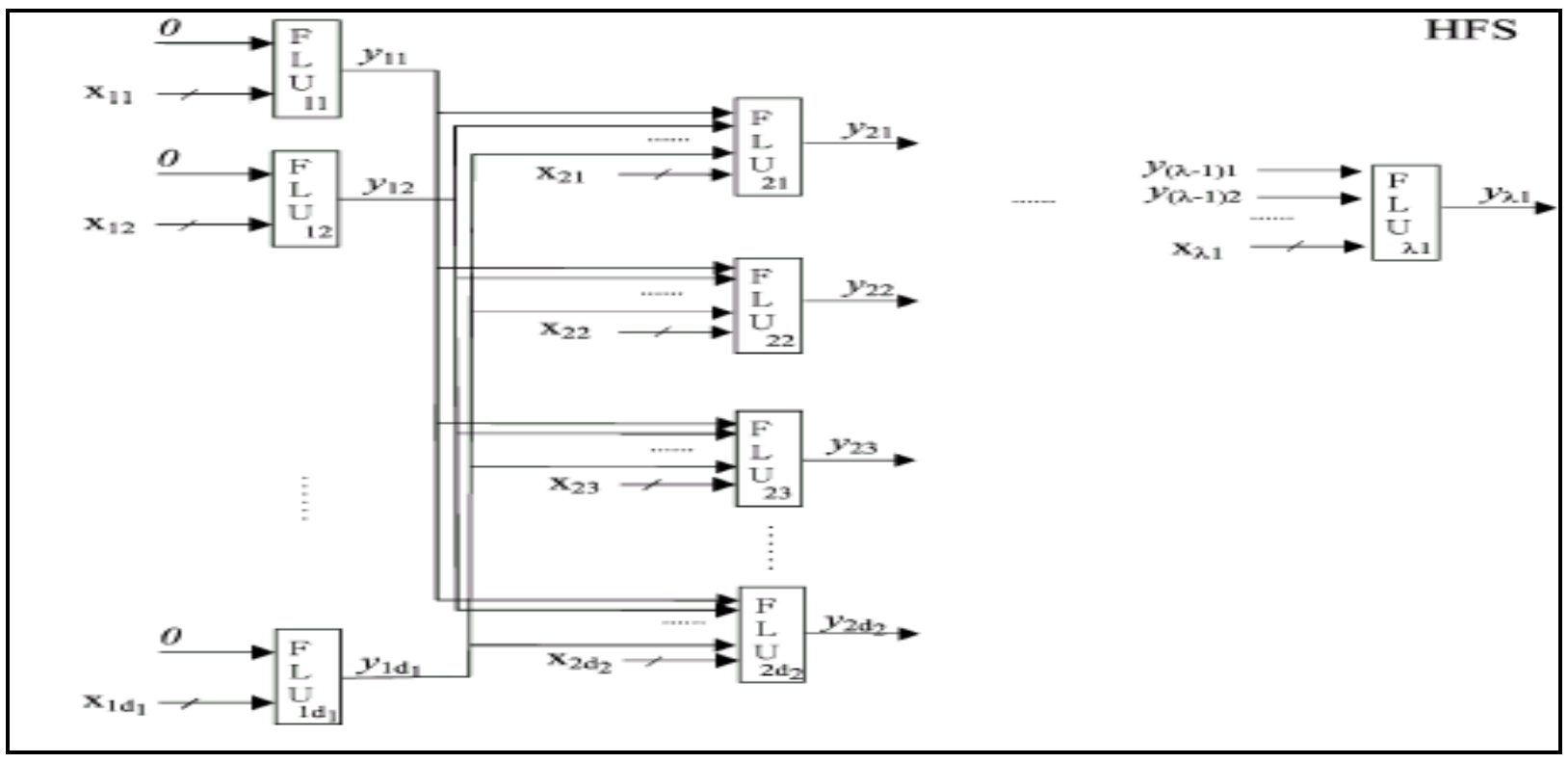

**Fig. 1.** Joo hierarchy fuzzy

This structure supports the use of all input at each level also Joo variables, uses the outputs of each FLU in the following by inserting them into every level FLÜS in levels that will follow.

## *2.3 Fuzzy hierarchy Jellali:*

The hierarchy proposed proceeds by use of two successive input vectors of inputs *X=(x1,x2,x3, ... xn)* T, two by two in a single FLU to converged property Wang said that the minimum basis of the rules is in the case of only uses two entries at each level [16]. For each pair of L level input is associated with two intermediate variables are Yi j such that i ∈ [1 .. L], j ∈ [1.. n]. These variables will be used in the conditional parts of the rules but the consequences of each one to which we retain the physical meaning of the rules therefore breaks in the system so that if n = variable number input pair is unlikely to have a remaining variable, otherwise we divided the variables in pairs and the remaining variable is added to the last level like the case of a typical decomposition presented by [17].

While referring to Wang, Hiewendiek Joo focuses on the formulation of the basic rules of combining the outputs of the previous not in the conditional parts of the rules, but in the consequences of the rules of the following levels to keep the physical meaning of the rules thing that is not obvious by conservation formulation of Wang. Nevertheless, the importance of approaching Joo, but it is complex because it combines the formulations of Wang and Brockmann[15].

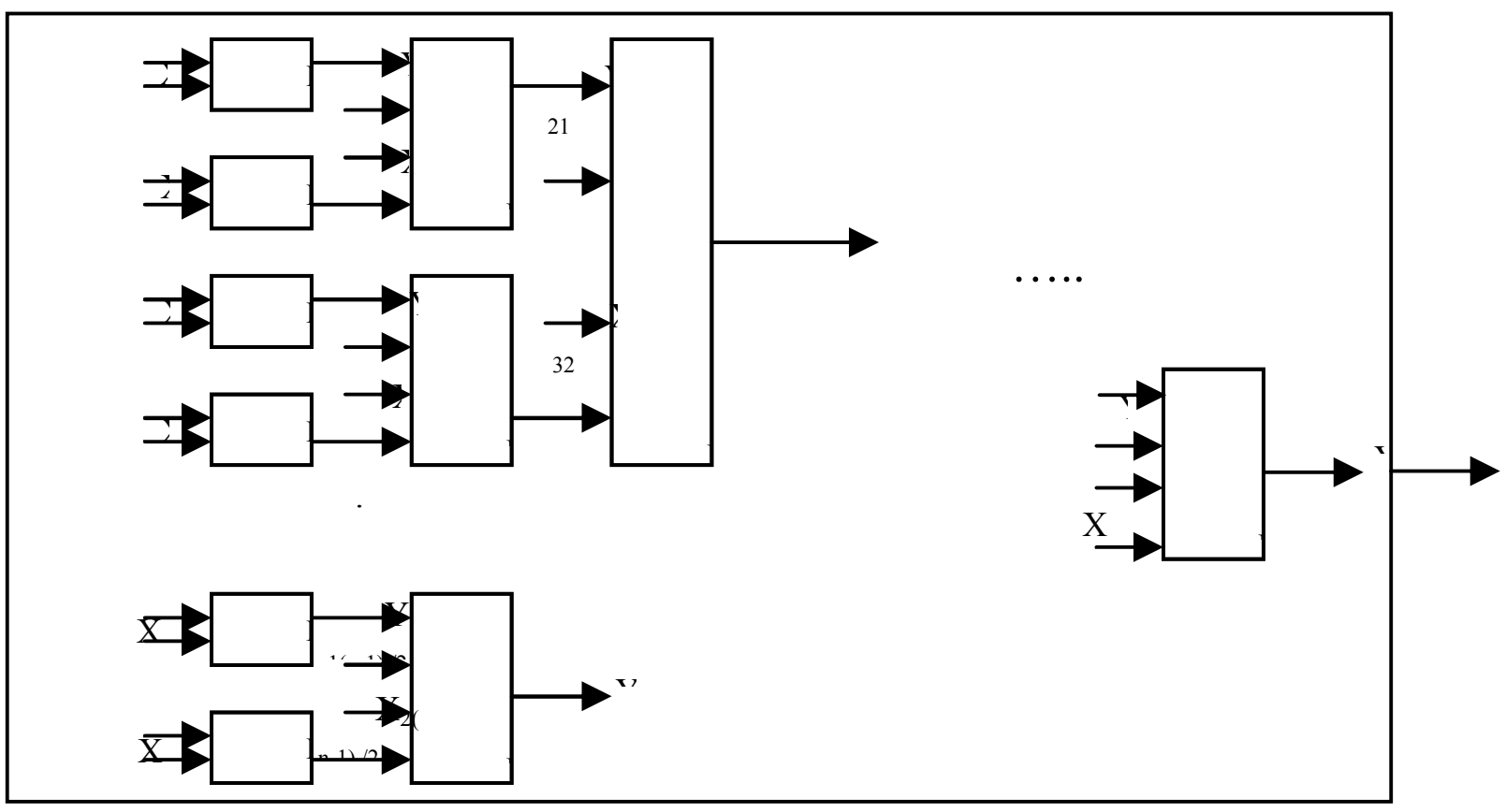

**Figure 2.** Jellali general hierarchy

Hierarchical fuzzy systems have created to address one of the major problems sudden to fuzzy systems standard multiple variables which is actually the exponential progress in the number of rules based on the number of variables entered in the system which leads to the impractical problem for the purposes of its implementation for the dimension of the basic rules.

## 3 Biped robot model

We considered a planar biped that has seven degrees of freedom and was based on five links (see Fig. 3). The model of the biped robot is presented at [16]. It is planned and can acts in the x-y plane. The structure was simple and consists on torso and knees without ankles and can represent simplified walking process.

The model of the biped robot that will be used is a walking robot with a torso and knees, without ankles. The tread surface can be seen as a sequence of points connected with straight lines. The two-dimensional biped consists of 5 segments that are connected with smooth joints.

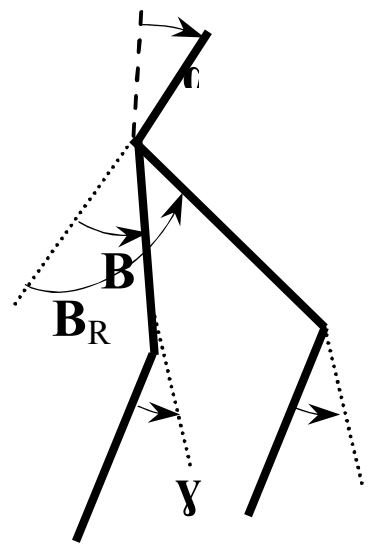

**Fig. 3.** Structure of the biped robot

## 4 Hierarchical fuzzy controller

In the literature, there are different concepts of logic and fuzzy control [10-11]. To design a fuzzy system with a fair amount of accuracy, increasing the number of input variables to the fuzzy system results in an exponential increase in the number of rules required [9]. Hierarchical fuzzy systems have created to address one of the major sudden problems to fuzzy multiple variables standard systems which is actually the exponential increase in the number of rules based on the number of variables entered in the system which leads to the impractical problem of the purposes of its implementation for the dimension of the basic rules[15]. The main challenge of our research is to propose an intelligent architecture and controller that are "humanly" inspired [13]. We propose the following structure of hierarchical fuzzy controller (Fig. 4 ):

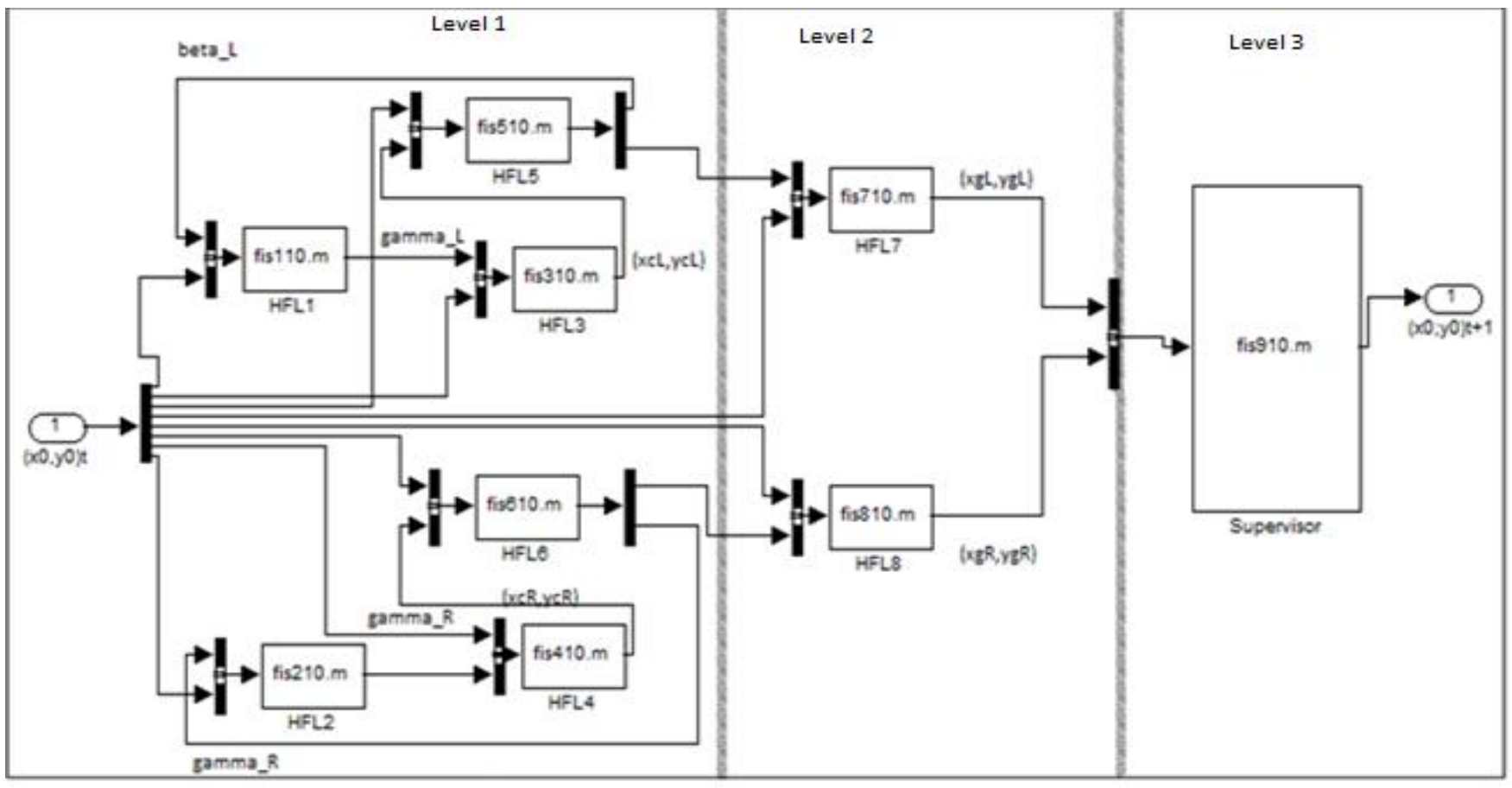

**Fig. 4.** General structure of the HFLC

Three hierarchical levels are presented on the robot: the level of the control Ankles (level 1), level of control (Level 2) knees and the level of control of the center of mass (level 3). The control is performed while minimizing the error between the planned trajectory and path actually followed. Therefore, the proposed controller may be formed by a parallel combination of four sub-controllers under the supervision of a controller. To achieve these controllers, we made the simulation of the biped robot model and all vectors that are removed from the robot states that will be taken as inputs and outputs for our different proposed controllers.

The controllers are based on neuro_fuzzy systems[13, 14]. We will therefore present the controllers HFL1, HFL3, HFL5 and HFL7. The latter is considered as a supervisor controller for this leg. For HFL1, the controller inputs are the vector (x0, y0) coordinates of the center of mass and the angle angel beta_left, the output is the same leg angle gamma_left. For learning process, we takes several vectors of different sizes (10, 30, 40, 60, 120 elements) that correspond to the inputs and outputs of the controller. The behavior of the hierarchical fuzzy controllers are represented in figure 5.

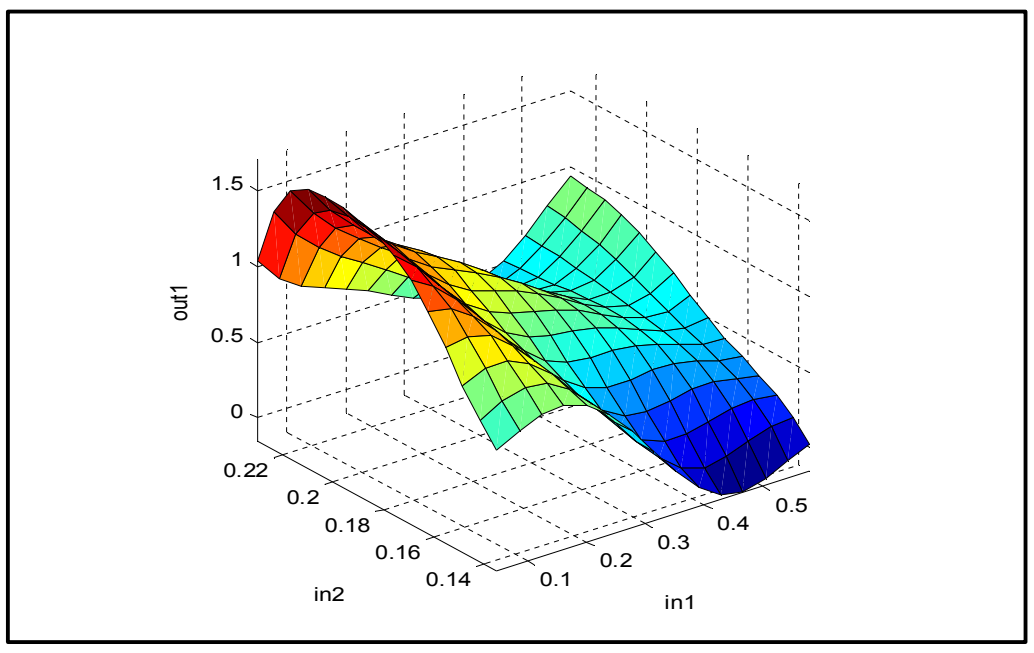

**Fig. 5.** Rules Surface viewer of HLF1

Hierarchical FLC has overcome the curse of dimensionality problem of fuzzy system and helped to reduce the complexity of the model due to the reduction rules used. The procedure for controllers testing is described as follows: we take a test vectors that correspond to the vectors of inputs and outputs of the controller.
To evaluate the proposed hierarchical fuzzy controller, we used the sum of squared errors from the test vector. The error of the controller, which is the value of the tested output subtracted from the actual output controller. For each controller, we calculate the index of the cumulative squared error and it's present according to the different values of training vectors. Fig. 6, shows the evolution of the index errors based on the basis of learning.

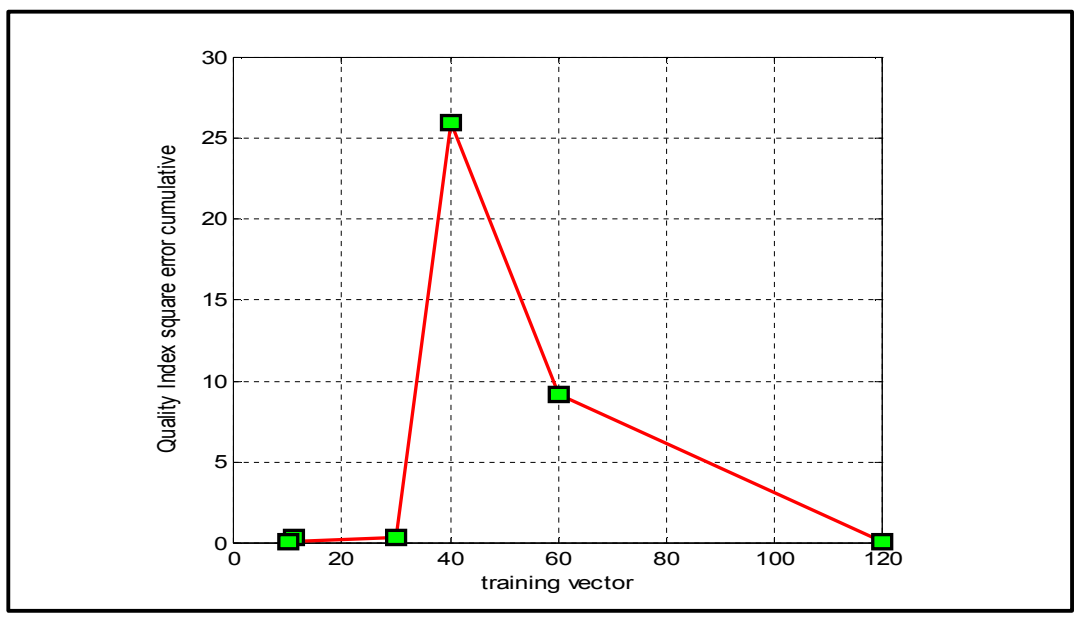

**Fig. 6.** Representation of the error of the proposed HFL1

For learning basic of 30 elements, the controller has not learned well enough which causes a significant error. For HFL3, the controller inputs are the vector (x0, y0) coordinates of the center of mass and the angle angel gamma_left, the output is the vector (xcl,ycl) coordinates of left ankle. For learning process, we takes several vectors of different sizes (10, 30, 40, 60, 120 elements) that correspond to the inputs and outputs of the controller. The behavior of the hierarchical fuzzy controllers is represented in figure 7.

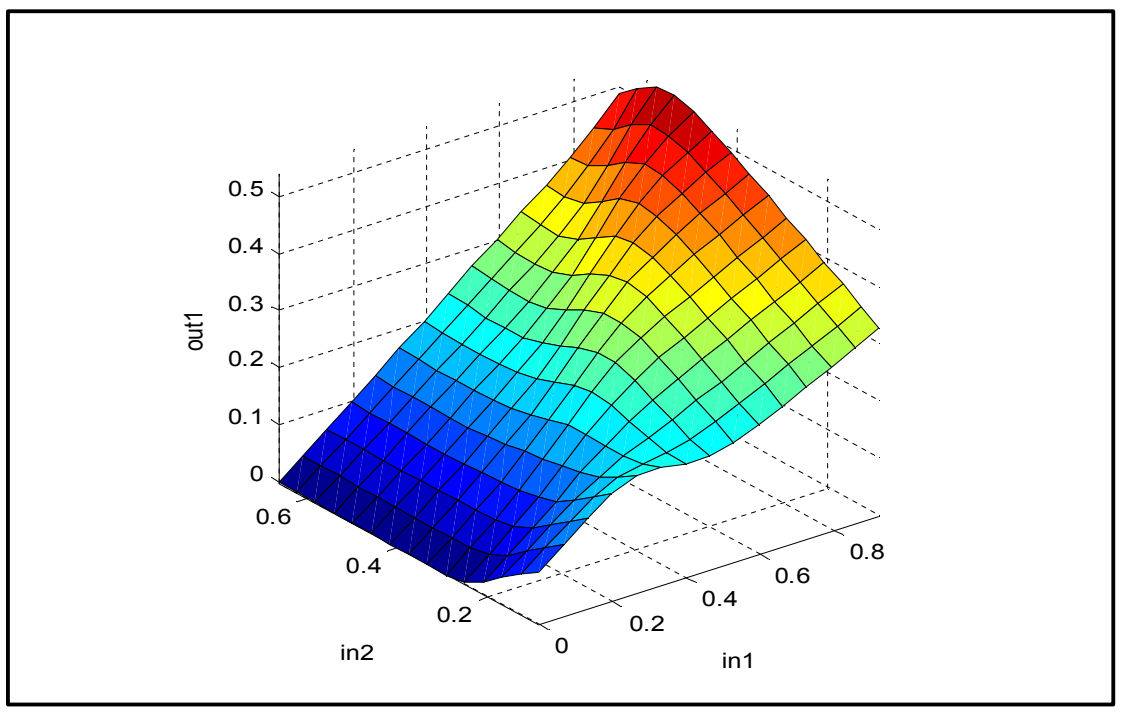

**Fig. 7.** Rules Surface viewer of HLF3

The following figure (Fig. 8) shows the evolution of the index errors based on the basis of learning.

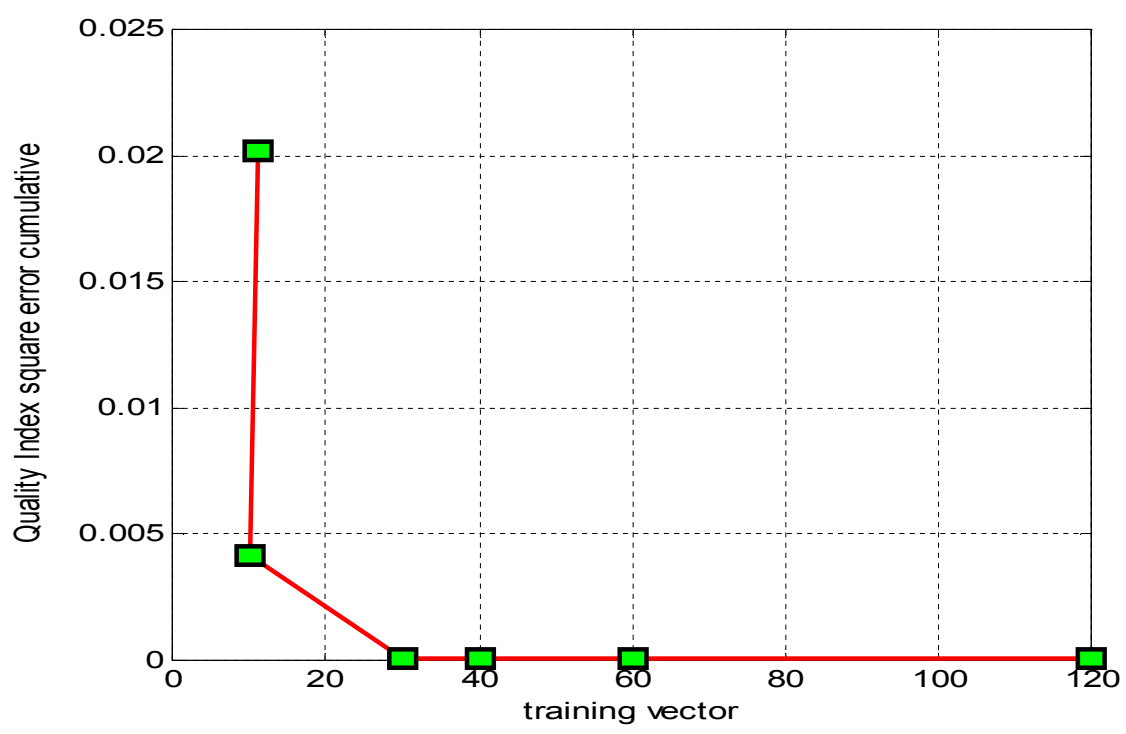

**Fig. 8.** Representation of the error of HFL3

The controller has learned the basics for all learning. The error is practically zero. For HFL5, the controller inputs are the vector (x0, y0) coordinates of the center of mass the vector (xcl,ycl) coordinates of left ankle.The output is the angel beta_left. For learning process, we takes several vectors of different sizes (10, 30, 40, 60, 120 elements) that correspond to the inputs and outputs of the controller. The behavior of the hierarchical fuzzy controllers are represented in (Fig. 9).

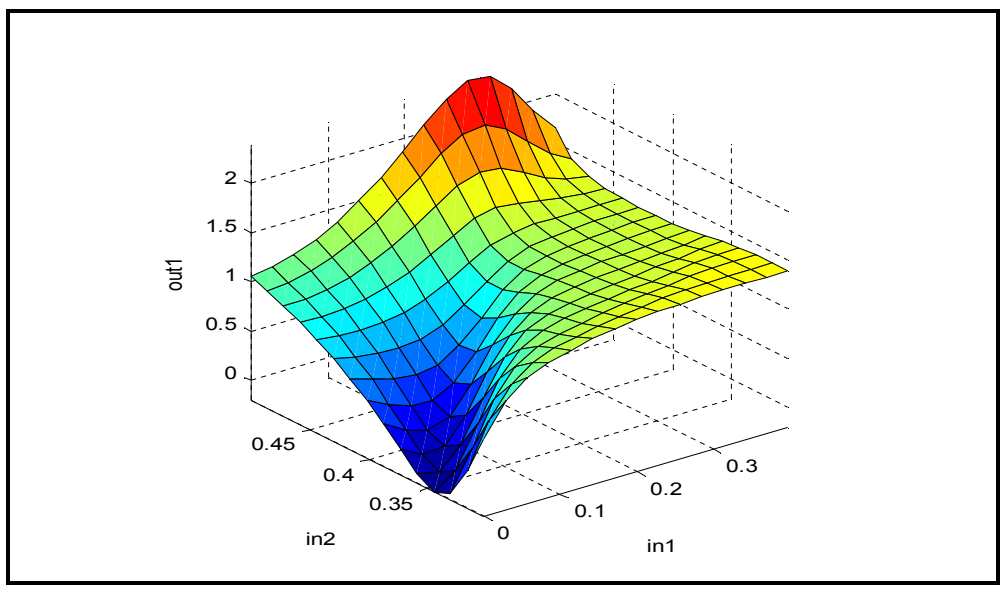

**Fig. 9.** Rules Surface viewer of HLF5

The error is calculated in the same way as HFL1 and HFL3. Figure 10 shows the evolution of the index errors based on the basis of learning.

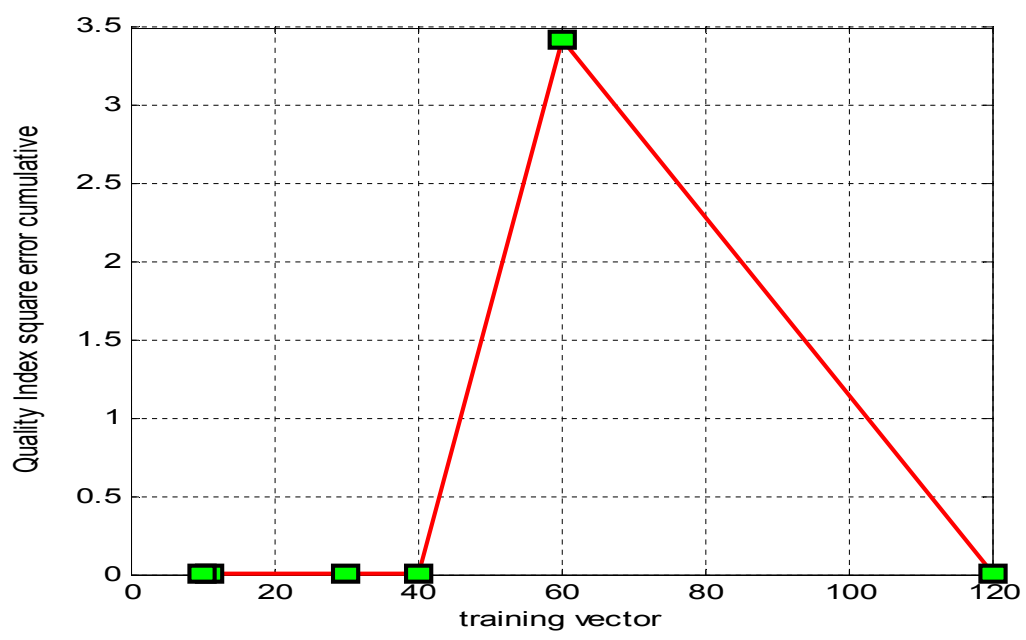

**Fig. 10.** Representation of the error of HFL5

For learning base with 60 elements, the controller does not behave well which gives a remarkable error. Now, we present the supervisor controller for the left ankle the HFL7 controller, which we consider as the supervisor controller of the left leg.

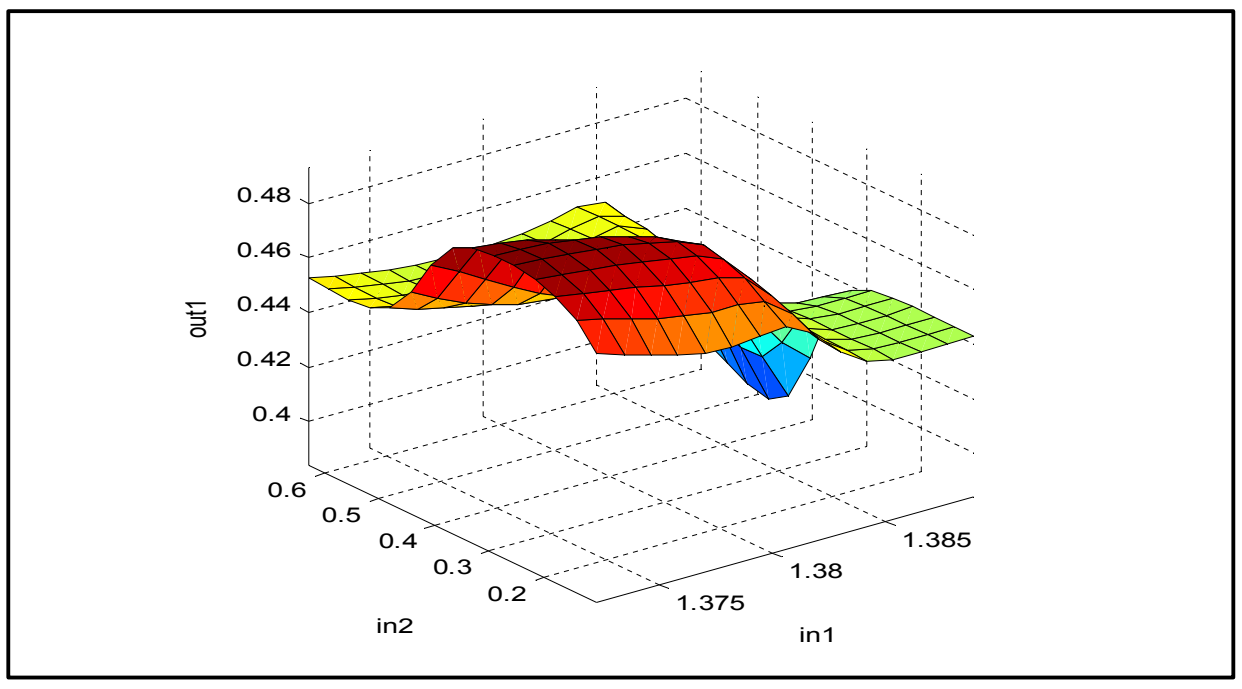

**Fig. 11.** Rules Surface viewer of HLF7

The controller inputs are the vector (x0, y0) coordinates of the center of mass and the angel beta_left. the output is the vector (xgl,ygl) coordinates of left knee. For learning process, we takes several vectors of different sizes (10, 30, 40, 60, 120) elements that correspond to the inputs and outputs of the controller. The behavior of the hierarchical fuzzy controllers was represented in the figure 11. The error is calculated in the same way as HFL1. Fig. 12 shows the evolution of the index errors based on the basis of learning.

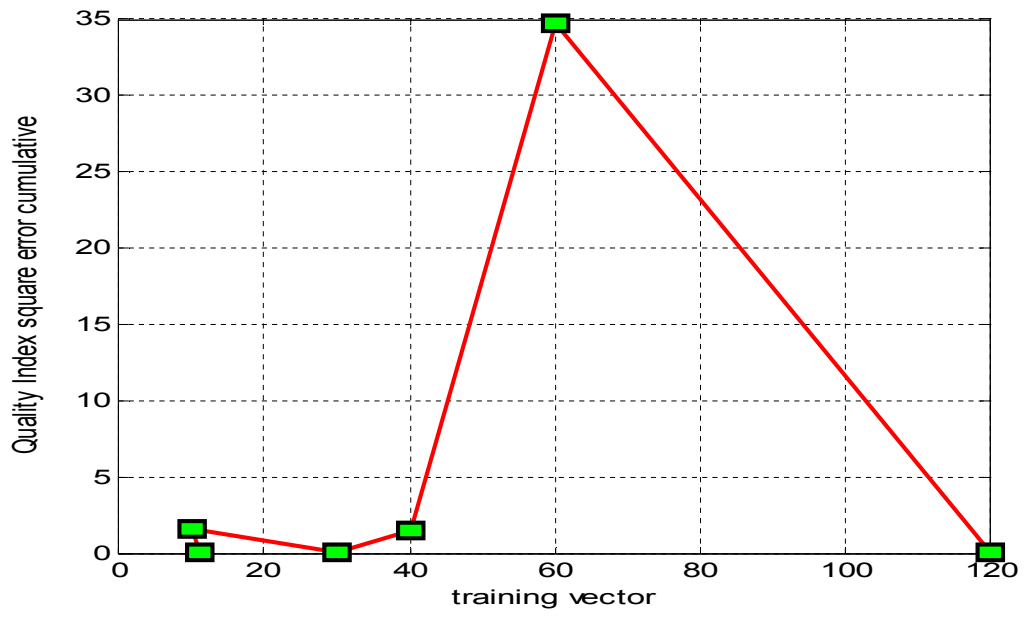

**Fig. 12**. Representation of the error of HFL7

## 5 Discussion and Conclusion

In our work, some problems of control and stability have not been taken into account: the contact between the foot and the ground is unilateral and can be broken in the presence of disturbances; the concept of stability must be properly adapted to the walking robot whose task is to make walk cycles.

In this work, the hierarchical structure of the biped robot is respected. Variable (x0, y0) is an input variable for all sub controllers. Our controller is based on the ANFIS adaptive system that is used for learning of the different controllers. Thus, we can easily check the stability of the robot during the walking phase.

By analyzing the different results of different hierarchical fuzzy controllers, we can see that;

- For a quick and efficient learning, we have used the expertise
- Have an effective but long learning, which corresponds to a sampling theorem of Shannon is used.

What can be observed, the symmetry between the two controllers of left and right legs.

The learning process has been considered in two ways:

- The first is that typically vague recourse expertises a biomechanics to adjust the selection of a relevant set of training and minimal
- A lack expertise, the controller is generated with maximized data (tens of minutes for training), but the error is respectable.

The recourse to an alternative learning proves to be a risky business as well as regards the nature of the controller that generated the measured error.

In conventional controllers based on fuzzy logic, the calculation complexity increases with the dimensions of the variables system. The proposed hierarchical fuzzy controller (HFLC) was implemented to reduce the number of rules to a lin-ear function of variables system. However, the use of hierarchical fuzzy logic controllers raises new issues in the automatic design of controllers, namely the coordination of outputs of sub-controllers at lower levels of the hierarchy.